\pdfoutput=1
\documentclass[10pt,twocolumn,letterpaper]{article}

\usepackage{iccv}
\usepackage{times}
\usepackage{epsfig}
\usepackage{graphicx}
\usepackage{amsmath}
\usepackage{amssymb}
\usepackage{times}
\usepackage{epsfig}
\usepackage{graphicx}
\usepackage{amsmath}
\usepackage{amssymb}
\usepackage{multirow}
\usepackage{booktabs} 
\usepackage{algorithm}
\usepackage{algorithmic}
\usepackage{subfigure}

\usepackage{color}

\usepackage[pagebackref=true,breaklinks=true,letterpaper=true,colorlinks,bookmarks=false]{hyperref}

\iccvfinalcopy 

\def\iccvPaperID{157} 

\ificcvfinal\fi
\begin{document}
	
	\title{Dual Motion GAN for Future-Flow Embedded Video Prediction}
	
	\author{Xiaodan Liang, Lisa Lee\\
		Carnegie Mellon University\\
		{\tt\small \{xiaodan1,lslee\}@cs.cmu.edu}
		\and
		Wei Dai, Eric P. Xing\\
		Petuum Inc.\\
		{\tt\small \{wei.dai,eric.xing\}@petuum.com}
	}
	
	\maketitle

	\begin{abstract}
		Future frame prediction in videos is a promising avenue for unsupervised video representation learning. Video frames are naturally generated by the inherent pixel flows from preceding frames based on the appearance and motion dynamics in the video. However, existing methods focus on directly hallucinating pixel values, resulting in blurry predictions. In this paper, we develop a dual motion Generative Adversarial Net (GAN) architecture, which learns to explicitly enforce future-frame predictions to be consistent with the pixel-wise flows in the video through a dual-learning mechanism. The primal future-frame prediction and dual future-flow prediction form a closed loop, generating informative feedback signals to each other for better video prediction. To make both synthesized future frames and flows indistinguishable from reality, a dual adversarial training method is proposed to ensure that the future-flow prediction is able to help infer realistic future-frames, while the future-frame prediction in turn leads to realistic optical flows. Our dual motion GAN also handles natural motion uncertainty in different pixel locations with a new probabilistic motion encoder, which is based on variational autoencoders. Extensive experiments demonstrate that the proposed dual motion GAN significantly outperforms state-of-the-art approaches on synthesizing new video frames and  predicting future flows. Our model generalizes well across diverse visual scenes and shows superiority in unsupervised video representation learning.
	\end{abstract}
	
	\section{Introduction}
	
	Despite the great progress of deep learning architectures for supervised learning,
	unsupervised video representation learning for general and scalable visual tasks remains a largely unsolved yet critical research problem. Recently, predicting future frames~\cite{oh2015action,mathieu2015deep,srivastava2015unsupervised} in a video sequence has surged as a promising direction for unsupervised learning of video data. 
	
	Video frame prediction itself is a challenging task due to the complex appearance and motion dynamics of natural scenes. Intuitively, in order to predict realistic pixel values in future frames, the model must be capable of capturing pixel-wise appearance and motion changes so as to let pixel values in previous frames flow into new frames. However, most existing state-of-the-art approaches~\cite{mathieu2015deep,srivastava2015unsupervised,lotter2016deep,liu2017video, sedaghat2016next,xue2016visual} use generative neural networks to directly synthesize RGB pixel values of future video frames and do not explicitly model the inherent pixel-wise motion trajectories, leading to blurry predictions. Although several recent attempts~\cite{patraucean2015spatio,liu2017video, sedaghat2016next} have tried to alleviate this issue by designing a motion field layer that copies pixels from previous frames, the predictions suffer from notable artifacts due to imprecise intermediate flows.
	
	In this work, we develop a dual motion Generative Adversarial Network (GAN) architecture that learns to explicitly make the synthesized pixel values in future frames coherent with pixel-wise motion trajectories using a dual adversarial learning mechanism. Specifically, it simultaneously resolves the primal future-frame prediction and dual future-flow prediction based on a shared probabilistic motion encoder. Inspired by the success of GANs~\cite{goodfellow2014generative,liang2017recurrent}, we establish a dual adversarial training mechanism between two future-frame and future-flow generators, and two frame and flow discriminators, to make the predictions indistinguishable from real data. The underlying dual-learning mechanism bridges the communication between future pixel hallucination and flow prediction by mutually reviewing each other. Our dual motion GAN consists of three fully-differentiable modules as follows.

	
	\begin{itemize}
		\vspace{-3mm}
		\item A probabilistic motion encoder captures motion uncertainty that may appear in different locations, and produces latent motion representations for preceding frames which are then fed as inputs to two generators.
		\vspace{-2mm}
		\item The future-frame generator then predicts future frames, which are assessed from two aspects: frame fidelity by a frame discriminator, and flow fidelity by passing the estimated flows between the preceding frame and the predicted frame into a flow discriminator.
		\vspace{-2mm}
		\item The future-flow generator in turn predicts future flows, which are also assessed from two aspects: flow fidelity by a flow discriminator, and frame fidelity by passing the extrapolated future frame (computed by a nested flow-warping layer) into a frame discriminator. 
		\vspace{-2mm}
	\end{itemize}

	By learning over symmetric feedback signals from two dual adversarial discriminators, the future-frame generator and future-flow generator mutually benefit from each other's complementary targets, leading to better video prediction. 
	Our dual motion GAN outperforms all existing approaches on synthesizing next frames and long-term future frames of natural scenes after training on car-mounted camera videos from the KITTI dataset~\cite{geiger2013vision} and consumer videos from the UCF-101 dataset~\cite{soomro2012ucf101}. We also demonstrate its generalization capability by testing on a different car-cam Caltech dataset~\cite{dollar2009pedestrian} and a collection of raw dash-cam videos from YouTube. In addition, we demonstrate the critical design choices of each module through extensive ablation studies. Further experiments on flow estimation, flow prediction, and action classification show our model's superiority on unsupervised video representation learning.

		\begin{figure*}[!tp]
			\begin{center}
				\includegraphics[scale=0.55]{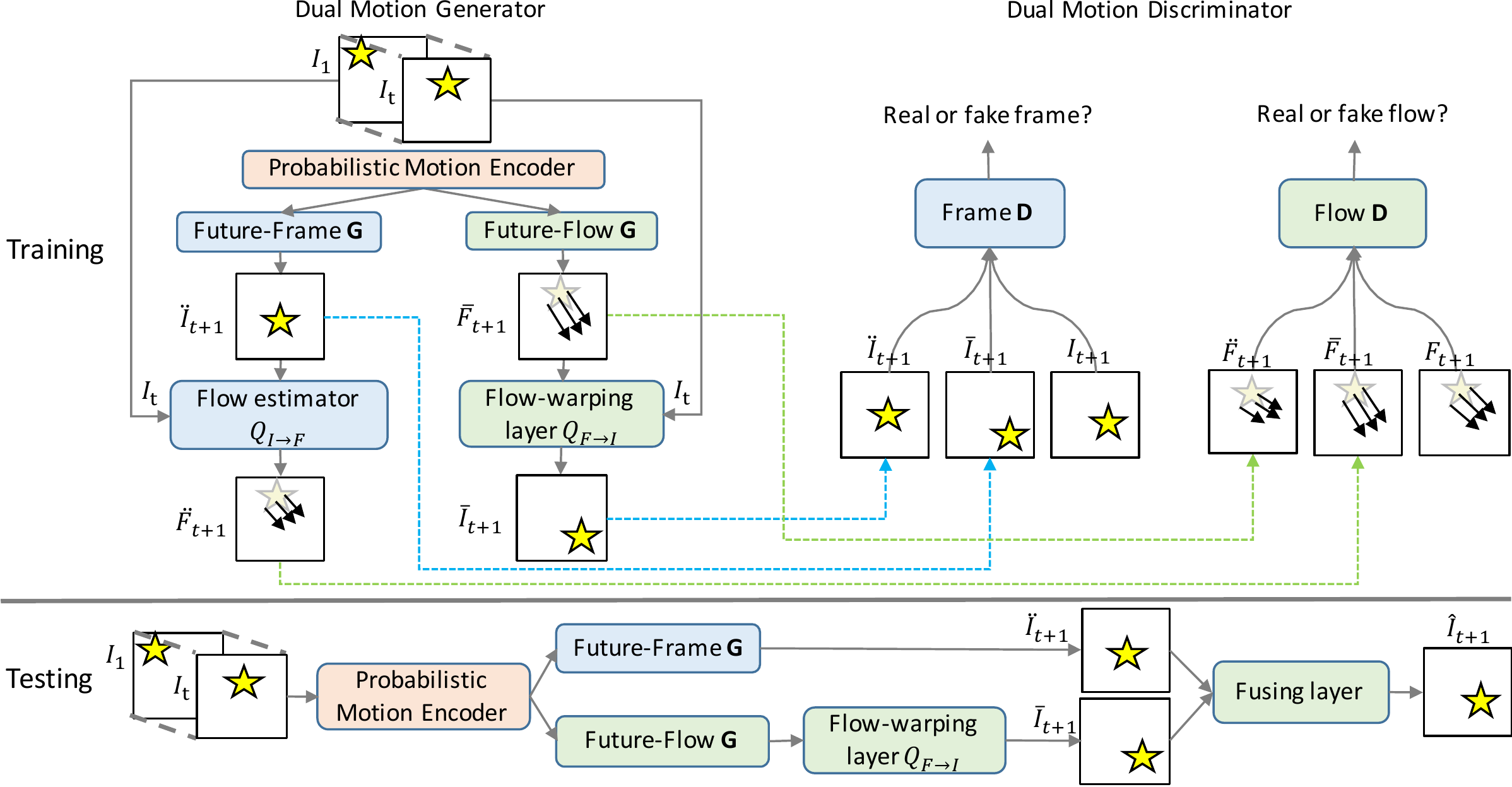}
				\caption{The proposed dual motion GAN jointly solves the future-frame prediction and future-flow prediction tasks with a dual adversarial learning mechanism. A video sequence $I_1, \dots, I_t$ is first fed into a probabilistic motion encoder $E$ to obtain a latent representation $z$. The dual motion generators (``Future-frame G'' and ``Future-flow G'' on the left) decode $z$ to synthesize future frames and flows. The dual motion discriminators (``Frame D'' and ``Flow D'' on the right) learn to classify between real and synthesized frames or flows, respectively. The flow estimator $Q_{I\rightarrow F}$ takes the predicted frame $\ddot{I}_{t+1}$ and real frame $I_t$ to estimate the flow $\ddot{F}_{t+1}$, which is further judged by ``Flow D''. The flow-warping layer $Q_{F \rightarrow I}$ warps the real frame $I_t$ with the predicted flow $\bar{F}_{t+1}$ to generate the warped frame $\bar{I}_{t+1}$, which is then evaluated by ``Frame D''. The testing stage is shown in the bottom row.} 
				\label{fig:framework}
			\end{center}
			\vspace{-9mm}
		\end{figure*}
	\section{Related Work}
	
	The proposed dual motion GAN attempts to jointly resolve the future-frame and future-flow prediction problems with a unified architecture. Our review thus focuses on two lines of literature most relevant to our problem.
	
	\textbf{Motion Prediction.} Various methods have been investigated to predict a future motion field~\cite{liu2011sift,walker2014patch,walker2016uncertain, long2016learning, LuoPHAF17} and visual representation~\cite{srivastava2015unsupervised,vondrick2016anticipating} given an image or a video sequence. Optical flow is the most commonly explored motion field, though large and fast motions can pose problems. Beyond deterministic motion prediction~\cite{liu2011sift, long2016learning}, a more recent work~\cite{walker2016uncertain} proposed using a variational autoencoder as a probabilistic prediction framework to handle intrinsic motion ambiguities. In contrast to prior works that only aim to generate optical flows, our dual motion GAN treats future-flow prediction as a dual task of future-frame prediction, and reviews the flow predictions by a frame discriminator using an dual adversarial learning mechanism. In addition, we introduce a novel probabilistic motion encoder that captures pixel-wise motion uncertainties to model long-term motion dynamics for boosting flow prediction.
	
	\textbf{Video Frame Prediction.} Many experiments with synthesizing video frames have been conducted recently~\cite{mathieu2015deep,lotter2016deep,liu2017video, sedaghat2016next,xue2016visual,patraucean2015spatio}. A line of research~\cite{mathieu2015deep,lotter2016deep, xue2016visual,vondrick2016generating} focuses on developing advanced networks to directly generate pixel values. However, they often produce blurry predictions since it is hard to model the complex pixel-level distributions of natural images. 
	Several approaches~\cite{patraucean2015spatio,liu2017video, sedaghat2016next} alleviate this blurring problem by resorting to motion field prediction for copying pixels from previous frames. 
 More recently, Sedaghat et al.~\cite{sedaghat2016next} explores a hybrid multi-task framework to jointly optimize optical flow estimation and frame prediction. However, these models still suffer from notable artifacts due to imprecise intermediate flows and unrealistic frame predictions. In contrast, our dual motion GAN learns to mutually optimize the future-frame generator and future-flow generator. It effectively alleviates the problem of frame deviations accumulating over time by incorporating pixel-wise motion trajectory prediction. Moreover, the dual frame and flow discriminators help drag the distributions of generated frames and flows closer to the real data distribution. Our model is thus able to produce sharp future frames and reasonable future flows simultaneously for a wide range of videos.
	
	\section{Dual Motion GAN}
	\label{sec:model}
	
\begin{figure*}[!tp]
			\begin{center}
				\includegraphics[scale=0.5]{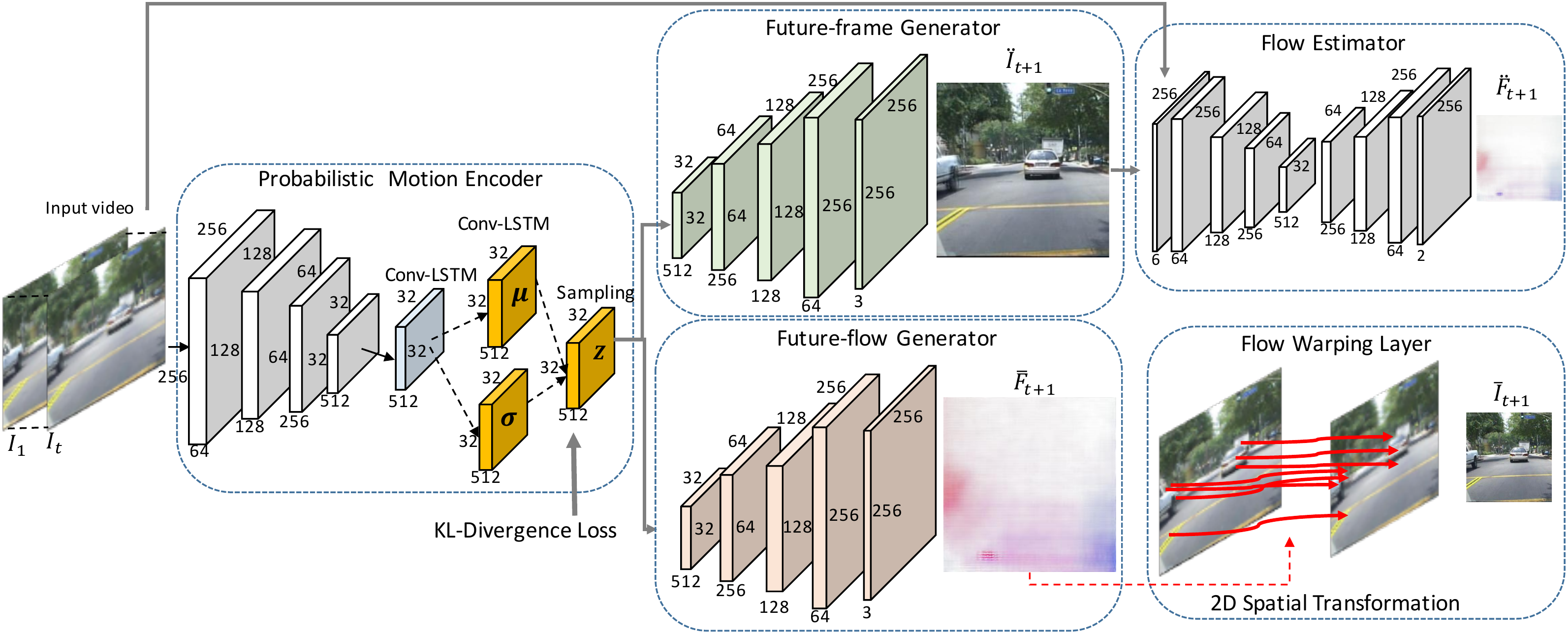}
				\caption{The dual motion generator. Each frame in a given sequence is recurrently fed into the probabilistic motion encoder $E$, which includes four convolutional layers, one intermediate ConvLSTM layer, and two ConvLSTM layers that produce the mean maps and variance maps for sampling $z$. Next, the future-frame generator $G_I$ and future-flow generator $G_F$ decode $z$ to produce a future-frame $\ddot{I}_{t+1}$ and future flow $\bar{F}_{t+1}$, respectively. The flow estimator $Q_{I\rightarrow F}$ then generates the estimated flow $\ddot{F}_{t+1}$ between $I_t$ and  $\ddot{I}_{t+1}$. The flow-warping layer $Q_{F\rightarrow I}$, which performs differential 2D spatial transformation, warps $I_t$ into $\bar{I}_{t+1}$ according to $\bar{F}_{t+1}$.} 
				\label{fig:network}
			\end{center}
			\vspace{-7mm}
		\end{figure*}

	We propose the dual motion GAN, a fully differentiable network architecture for video prediction that jointly solves the primal future-frame prediction and dual future-flow prediction. The dual motion GAN architecture is shown in Figure~\ref{fig:framework}. Formally, our dual motion GAN takes a video sequence $\mathbf{v} = \{I_1, \dots, I_t\}$ as input to predict the next frame $\hat{I}_{t+1}$ by fusing the future-frame prediction $\ddot{I}_{t+1}$ and future-flow based prediction $\bar{I}_{t+1}$. We adopt the simple $1\times1$ convolution filters for the fusing operation. The dual motion generator (shown in Figure~\ref{fig:network}) consists of five components: a probabilistic motion encoder $E$, future-frame generator $G_I$, future-flow generator $G_F$, flow estimator $Q_{I\rightarrow F}$ and flow-warping layer $Q_{F\rightarrow I}$. The dual motion discriminator (shown in Figure~\ref{fig:discriminator}) consists of a frame discriminator $D_I$ and a flow discriminator $D_F$. More specifically, the probabilistic motion encoder $E$ first maps previous frames to a latent code $z$. The future-frame generator $G_I$ and future-flow estimator $G_F$ then decode $z$ to predict the future frame $\ddot{I}_{t+1}$ and future flow $\bar{F}_{t+1}$, respectively. The fidelity of $\ddot{I}_{t+1}$ is judged by how well $\ddot{I}_{t+1}$ fools the frame discriminator $D_I$, and how well the flow $\ddot{F}_{t+1}$ between $I_t$ and $\ddot{I}_{t+1}$, estimated using the flow estimator $Q_{I\rightarrow F}$, fools the flow discriminator $D_F$. Similarly, the quality of future-flow prediction is judged by how well $\bar{F}_{t+1}$ fools the flow discriminator $D_F$, and how well the warped frame $\bar{I}_{t+1}$, generated by warping $I_t$ with $\bar{F}_{t+1}$ using the flow-warping layer $Q_{F\rightarrow I}$, fools the frame discriminator $D_I$. %


	\subsection{Adversarial Dual Objective}
	\label{sec:objective}	
	In this section, we formally derive the training objective of our dual motion GAN. 
	
	\textbf{VAE:} The encoder-generator triplet $\{E, G_I, G_F\}$ constitutes a variational autoencoder (VAE). The probabilistic motion encoder $E$ first maps a video sequence $\mathbf{v}$ into a code $z$ in the latent space $\mathcal{Z}$, and $G_I,G_F$ then decode a randomly perturbed version of $z$ to predict future frames and flows, respectively. Following~\cite{kingma2013auto}, we assume the components in the latent space $\mathcal{Z}$ are conditionally independent and Gaussian. The encoder $E$ outputs the mean maps $E_\mu(\mathbf{v})$ and the variance maps $E_{\sigma^2}(\mathbf{v})$, where the distribution of the latent code $z$ is given by $q(z|\mathbf{v}) = \mathcal{N}(z|E_\mu(\mathbf{v}), E_{\sigma^2}(\mathbf{v}))$. The architecture of $E$ is detailed in Section~\ref{sec:encoder}. The frame prediction is obtained as $\ddot{I}_{t+1} = G_I(z\sim q(z|\mathbf{v}))$, and the corresponding estimated flow is $\ddot{F}_{t+1} = Q_{I\rightarrow F}(\ddot{I}_{t+1},\mathbf{v})$. The flow prediction is calculated as $\bar{F}_{t+1} = G_F(z\sim q(z|\mathbf{v}))$, and the corresponding warped frame is $\bar{I}_{t+1} = Q_{F\rightarrow I}(\bar{F}_{t+1},\mathbf{v})$.  Note that the flow-warping layer $Q_{F\rightarrow I}$ does not have parameters to be optimized. We train the VAE by minimizing a variational upper bound of a negative log-likelihood function:
	\begin{equation}
	\begin{split}
	&\mathcal{L}_{\text{VAE}}(E, G_I, G_F, Q_{I\rightarrow F})=\mathbb{E}_{z\sim q(z|\mathbf{v})}(\\
	&-\log p_{G_I}(I_{t+1}|z) -\log p_{Q_{I\rightarrow F}}(F_{t+1}|G_I(z))\\
	&-\log p_{G_F}(F_{t+1}|z) -\log p_{Q_{F\rightarrow I}}(I_{t+1}|G_F(z))\\
	&+ \mathbf{KL}(q(z|\mathbf{v})||p(z)),
	\label{VAE}
	\end{split}
	\end{equation}
	where KL is the Kullback-Leibler divergence that penalizes deviation of the distribution of the latent code from the prior distribution $p(z) = \mathcal{N}(z|0, I)$. L1 distance~\cite{lotter2016deep,liu2017video} is imposed on the future-frame prediction $\ddot{I}_{t+1}$ and warped frame prediction $\bar{I}_{t+1}$. We thus model the conditional distribution $p_{G_I}(I_{t+1}|z) = \exp(-||I_{t+1}-G_I(z)||_1)$ and $p_{Q_{F\rightarrow I}}(I_{t+1}|G_F(z)) =  \exp(-||I_{t+1}-Q_{F\rightarrow I}(G_F(z),\mathbf{v})||_1)$. Following common practice in flow estimation~\cite{ranjan2016optical}, we adopt the average End Point Error (EPE) $\Delta_{\text{EPE}}$ to optimize the future-flow prediction and flow estimation. We thus compute two conditional distributions of flows as $p_{G_F}(F_{t+1}|z) = \exp(-\Delta_{\text{EPE}}(F_{t+1}, G_F(z)))$ and $p_{Q_{I\rightarrow F}}(F_{t+1}|G_I(z)) = \exp(-\Delta_{\text{EPE}}(F_{t+1}, Q_{I\rightarrow F}(G_I(z),\mathbf{v})))$. Hence, minimizing the negative log-likelihood term is equivalent to minimizing L1 distance between the predicted frame and the true frame, and the EPE loss between the predicted and the true flow.

	
	\textbf{Adversarial Dual Objective: } The generators $G_I, G_F$ and the discriminators $D_I, D_F$ form two dual generative adversarial networks, and enables the dual motion GAN to generate sharper and more realistic frame and flow predictions. As discussed in the new Wasserstein GAN (WGAN)~\cite{arjovsky2017wasserstein}, the original GAN~\cite{goodfellow2014generative} suffers from several training difficulties such as mode collapse and instable convergence. We thus follow the  proposed training strategies in Wasserstein GAN (WGAN)~\cite{arjovsky2017wasserstein} that theoretically remedy these problems by minimizing an approximated Wasserstein distance. The dual motion GAN can be trained by jointly solving the learning problems of the VAE and two dual GANs:
	\begin{equation}
	\begin{split}
	\min_{E, G_I, G_F, Q_{I\rightarrow F}}&\max_{D_I, D_F} \mathcal{L}_{\text{VAE}}(E, G_I, G_F, Q_{I\rightarrow F})\\ &+\lambda\mathcal{L}_{\text{GAN}}^I(G_I, G_F, D_I)  \\&+ \lambda\mathcal{L}_{\text{GAN}}^F(G_F, G_I, D_F, Q_{I\rightarrow F}). 
	\label{global}
	\end{split}
	\end{equation}
	where $\lambda$ balances the VAE loss and two dual GAN losses. The generators $G_I, G_F$ and flow estimator $Q_{I\rightarrow F}$ try to minimize the whole objective against the adversarial discriminators $D_I$ and $D_F$ that try to maximize it. The discriminator $D_I$ learns to distinguish real frames $I_{t+1}$ from the predicted frames $\ddot{I}_{t+1}$ and warped frames $\bar{I}_{t+1}$. Similarly, $D_F$ learns to distinguish real flows $F_{t+1}$ from predicted flows $\bar{F}_{t+1}$ and estimated flows $\ddot{F}_{t+1}$. Let $p(\mathbf{v})$ and $p(\mathcal{F})$ denote the true data distributions of true frames and true flows. The dual GAN objective functions are given by:
	\begin{equation}
	\begin{split}
	\mathcal{L}_{\text{GAN}}^I(G_I, G_F, &D_I) = \mathbb{E}_{I_{t+1} \sim p(\mathbf{v})} D_I(I_{t+1})\\& - \frac{1}{2}\mathbb{E}_{z \sim q(z|\mathbf{v})} D_I(G_I(z)) \\&- \frac{1}{2}\mathbb{E}_{z \sim q(z|\mathbf{v})} D_I(Q_{F\rightarrow I}(G_F(z))),\\
	\mathcal{L}_{\text{GAN}}^F(G_F, G_I, &D_F, Q_{I\rightarrow F}) = \mathbb{E}_{F_{t+1} \sim p(\mathcal{F})} D_F(F_{t+1})\\& - \frac{1}{2}\mathbb{E}_{z \sim q(z|\mathbf{v})} D_F(G_F(z))\\& - \frac{1}{2}\mathbb{E}_{z \sim q(z|\mathbf{v})} D_F(Q_{I\rightarrow F}(G_I(z))).
	\end{split}
	\end{equation}
	Our adversarial dual objective functions differ from the standard GAN objective function in that the samples for each discriminator come from two different distributions depicted by two dual generators. For $\mathcal{L}_{\text{GAN}}^I$, two synthesized distributions are $p_{G_I}$ and $p_{Q_{F\rightarrow I}}$ functioning on the distribution $p_{G_F}$. Optimizing $\mathcal{L}_{\text{GAN}}^I$ encourages both $G_I$ and $Q_{F\rightarrow I}$ to output frames resembling true frames from $p(\mathbf{v})$, which can further serve as feedback signals to the distribution $p_{G_F}$. Similarly, optimizing $\mathcal{L}_{\text{GAN}}^F$ encourages both $G_F$ and $G_I$ to output flows resembling true flows from $p(\mathcal{F})$. During training, the dual motion GAN uses true frames $I_{t+1}$ in the video to supervise future-frame prediction, and the optical flows $F_{t+1}$ estimated by EpicFlow~\cite{revaud2015epicflow} to supervise future-flow prediction. We choose the traditional EpicFlow since it does not require annotated flows for training, yet achieves the best results on flow estimation.
	
	\textbf{Learning:} Inheriting from GAN, the optimization of dual motion GAN can be seen as a two-player game---the first player consisting of an encoder and two generators, and the second player consisting of two adversarial discriminators. The first player's objective is to defeat the second player and also to minimize the VAE losses. Following WGAN~\cite{arjovsky2017wasserstein}, we apply an alternating gradient update scheme, performing five gradient descent steps on $D_I$ and $D_F$, and one step on $G_I, G_F, Q_{I\rightarrow F}$. We use minibatch SGD and apply the RMSprop solver~\cite{tieleman2012lecture}. The $\lambda$ is empirically set to 0.001, and the learning rate is set to 0.0001. We train the model for roughly 40 epochs. In order to have parameters of $D_I$ and $D_F$ lie in a compact space, we clamp the weights to a fixed box [−0.01, 0.01] after each gradient update. We apply batch normalization~\cite{ioffe2015batch} and set the batch size to 1, which has been termed ``instance normalization'' and demonstrated to be effective in image generation tasks. 
	
	\subsection{Network Architectures}
	
\begin{figure}[!tp]
			\begin{center}
			\includegraphics[scale=0.45]{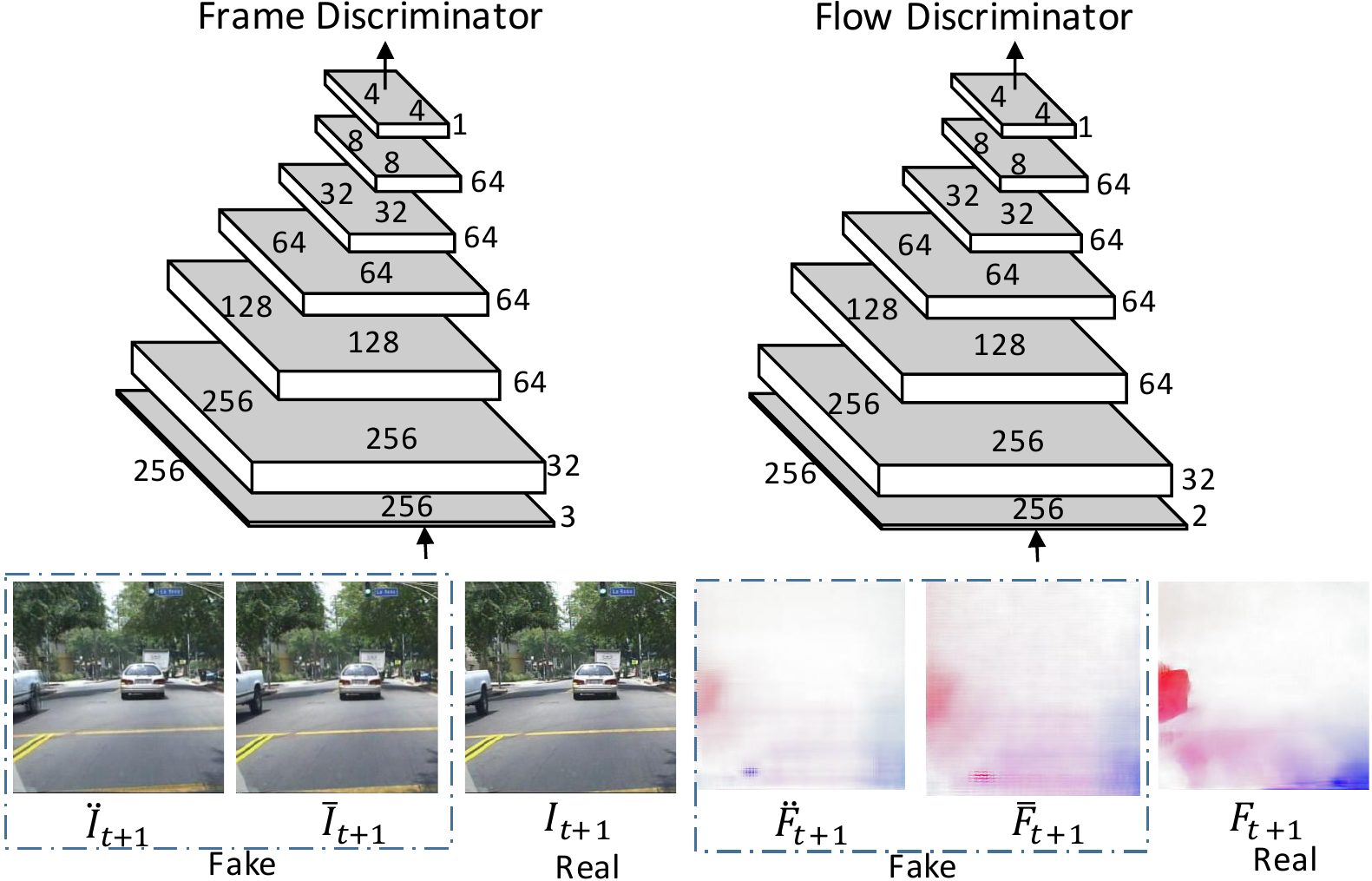}
				\caption{Architectures of the two dual motion discriminators. The frame and flow discriminators learn to classify between real and synthesized frames and flows, respectively.}
				\label{fig:discriminator}
			\end{center}
			\vspace{-8mm}
		\end{figure}

	The detailed networks for generators and discriminators are provided in Figure~\ref{fig:network} and Figure~\ref{fig:discriminator}, respectively. For simplicity, the pooling layers, batch normalization layers, and ReLU layers after the intermediate convolutional layers are omitted in the figures.
	
	\textbf{Probabilistic Motion Encoder:}
	\label{sec:encoder}
	The exact motions of objects in real-world videos are often unpredictable and have large variations due to intrinsic ambiguities. Existing works~\cite{xue2016visual,walker2016uncertain,kingma2013auto,goyal2017nonparametric} often learn a whole latent vector $z$ for all objects in the scene. A shortcoming of these models is that they cannot distinguish the particular motion pattern for each pixel location of distinct objects. We thus extend the variational autoencoder to generate a spatial joint distribution conditioned on the input frames. 
	
	Formally, to accommodate our dual motion GAN for an arbitrary number of input frames, we design a recurrent probabilistic motion encoder $E$ (Figure~\ref{fig:network}) to learn variational motion representations $z$ that encode past motion patterns and also model the uncertainty in motion fields. As presented in Section~\ref{sec:objective} (VAE), the encoder $E$ generates the mean maps $E_\mu(\mathbf{v})$ and the variance maps $E_{\sigma^2}(\mathbf{v})$. Specifically, each frame $I_t$ in $\mathbf{v}$ is recurrently passed into four convolutional layers to obtain a compact $32\times 32 $ feature map with 512 dimensions. Next, one Convolution LSTM (ConvLSTM) layer~\cite{hochreiter1997long,xingjian2015convolutional,liang2017interpretable} is employed for sequence modeling where the memory cells essentially act as an accumulator of the spatial state information. The resulting features map is further fed into two convolution LSTM layers to predict the mean maps $E_\mu(\mathbf{v})$ and variance maps $E_{\sigma^2}(\mathbf{v})$, respectively. Compared to conventional convolution layers, ConvLSTM determines the future state of a certain cell in the grid by incorporating the inputs and past states of its local neighbors. We use a $4\times4$ kernel and 512 hidden states for all ConvLSTM layers. Finally, the latent motion representation $z$ is sampled from $\mathcal{N}(z|E_\mu(\mathbf{v}), E_{\sigma^2}(\mathbf{v}))$.
	
	
	\textbf{Dual Motion Generator:}
	The future-frame generator $G_I$ decodes the shared motion representation $z$ to produce a future-frame prediction $\ddot{I}_{t+1}$ with RGB channels. Similarly, future-flow generator $G_F$ decodes $z$ to produce a future-flow prediction $\bar{F}_{t+1}$ with two channels that represent the horizontal and vertical components of the optical flow field. Both generators use five deconvolutional layers with $3 \times 3$ kernels. The flow estimator $Q_{I\rightarrow F}$, which consists of four convolutional layers and four deconvolutional layers, estimates flow maps $\ddot{F}_{t+1}$ between the previous frame $I_t$ and the future-frame prediction $\ddot{I}_{t+1}$. The flow-warping layer $Q_{F\rightarrow I}$ is a warp operator that generates $\bar{I}_{t+1}$ by warping the previous frame $I_t$ according to the predicted flow field $\bar{F}_{t+1}$ using bilinear interpolation. We follow the differential spatial transformation layer used in~\cite{ranjan2016optical,jaderberg2015spatial} to define this warping operator.
	
	\textbf{Dual Motion Discriminator:}
	As shown in Figure~\ref{fig:discriminator}, to optimize the adversarial dual objective in Eqn. (\ref{global}), the frame discriminator $D_I$ takes a $3\times256\times256$ image as input and produces one output value, while the flow discriminator $D_F$ takes a $2\times256\times256$ flow map from a real data or generated distribution. Following the Wasserstein GAN~\cite{arjovsky2017wasserstein}, we drop the Sigmoid operation in the final prediction in order to remedy the mode collapse problem in vanilla GANs~\cite{goodfellow2014generative}.

	\section{Experiments}
	In this section, we first present the main comparisons on video prediction tasks, including next frame and multiple frame predictions. We then present ablation studies on our model. In addition, we demonstrate the generalization capabilities of our model through extensive experiments on flow prediction, flow estimation, and unsupervised representation learning.
	
	\subsection{Comparisons on Video Prediction}
	\textbf{Experimental Settings.} We evaluate the video prediction capabilities of our model on complex real-world sequences. First, the performance is compared on the car-mounted camera videos. Following the state-of-the-art PredNet~\cite{lotter2016deep}, models are trained using raw videos from the KITTI dataset~\cite{geiger2013vision} and evaluated on the test partition of the Caltech Pedestrian dataset~\cite{dollar2009pedestrian}. Following PredNet's~\cite{lotter2016deep} procedure, 57 recording sessions from the “City”, “Residential”, and “Road” categories are used for training (roughly 41k frames) and 4 used for validation. In order to further validate our model's generalization capability, we evaluate the trained model on 500 raw 1-minute clips from YouTube, collected using the keywords ``dashboard videos''. Second, following~\cite{mathieu2015deep} and~\cite{liu2017video}, we train models on the generic consumer videos from UCF101~\cite{soomro2012ucf101}, and evaluate on the UCF-101~\cite{soomro2012ucf101} and THUMOS-15~\cite{gorban2015thumos} test sets. To compare with current state-of-the-art~\cite{liu2017video} models, we also use two previous frames as the input to predict the next future frame.
	
	We used the metrics MSE~\cite{lotter2016deep}, PSNR, and SSIM~\cite{wang2004image} to evaluate the image quality of video frame prediction, where higher values of PSNR and SSIM indicate better results. The implementations are based on the public Torch7 platform on a single NVIDIA GeForce GTX 1080. The details of our optimization procedure and hyperparameters are presented in Section~\ref{sec:model}. Our dual motion GAN takes around 300ms to predict one future frame and flow given a sequence of 10 previous frames (as in the Caltech test set).
	
	\begin{table}[!tp]
		\centering\renewcommand\arraystretch{1.3}\footnotesize
		\caption{Performance (MSE and SSIM) of video frame prediction on Caltech and YouTube clips after training on KITTI dataset.}\label{tab:kitti}
		\begin{tabular}{c|c|c|c|ccccccccccccccccccp{6em}p{5em}}
			\toprule
			& \multicolumn{2}{|c|}{Caltech} &\multicolumn{2}{|c}{YouTube Clip}\\
			\hline
			Method & MSE & SSIM  & MSE & SSIM\\
			\hline
			CopyLast & 0.00795 & 0.762 & 0.01521 & 0.785\\
			\hline
			BeyondMSE~\cite{mathieu2015deep} & {0.00326} & {0.881} & {0.00853} & {0.820}\\
			\hline
			PredNet~\cite{lotter2016deep} & 0.00313 & 0.884 & 0.00679 & 0.858\\
			\hline
			{Ours frame w/o GAN} & {0.00307} & {0.880} & {0.00833} & {0.826}\\
			{Ours frame GAN} & {0.00291} & {0.883} & {0.00793} & {0.836}\\
			{Ours flow w/o GAN} & {0.00292} & {0.884} & {0.00778} & {0.839}\\
			{Ours flow GAN} & {0.00289} & {0.887} & {0.00701} & {0.843}\\
			{Ours frame+flow w/o GAN} & {0.00269} & {0.892} & {0.00617} & {0.859}\\
			{Ours w/o motion encoder} & {0.00262} & {0.895} & {0.00583} & {0.863}\\
			\hline
			{Ours future-flow (testing)} & {0.00255} & {0.896} & {0.00601} & {0.866}\\
			{Ours future-frame (testing)} & {0.00260} & {0.893} & {0.00613} & {0.862}\\
			\hline
			\textbf{Ours (full)} & \textbf{0.00241} & \textbf{0.899} & \textbf{0.00558} & \textbf{0.870}\\
			\bottomrule
		\end{tabular}%
		\vspace{-3mm}
	\end{table}%
	
	\begin{table}[!tp]
		\centering\renewcommand\arraystretch{1.3}\footnotesize
		\caption{Performance (PSNR and SSIM) of video frame prediction on UCF-101 and THUMOS-15.}\label{tab:ucf}
		\begin{tabular}{c|c|c|c|ccccccccccccccccccp{6em}p{5em}}
			\toprule
			& \multicolumn{2}{|c|}{UCF-101} &\multicolumn{2}{|c}{THUMOS-15}\\
			\hline
			Method & PSNR & SSIM  & PSNR & SSIM\\
			\hline
			BeyondMSE~\cite{mathieu2015deep} & {28.2} & {0.89} & {27.8} & {0.87}\\
			\hline
			EpicFlow~\cite{revaud2015epicflow} & {29.1} & {0.91} & {28.6} & {0.89}\\
			\hline
			DVF~\cite{liu2017video} & 29.6 & 0.92 & 29.3 & 0.91\\
			\hline
			Nextflow~\cite{sedaghat2016next} & 29.9 & - & - & - \\
			\hline
			\textbf{Ours (full)} & \textbf{30.5} & \textbf{0.94} & \textbf{30.1} & \textbf{0.92}\\
			\bottomrule
		\end{tabular}%
	\end{table}%
	
	\begin{figure}[!tp]
		\begin{center}
			\includegraphics[scale=0.32]{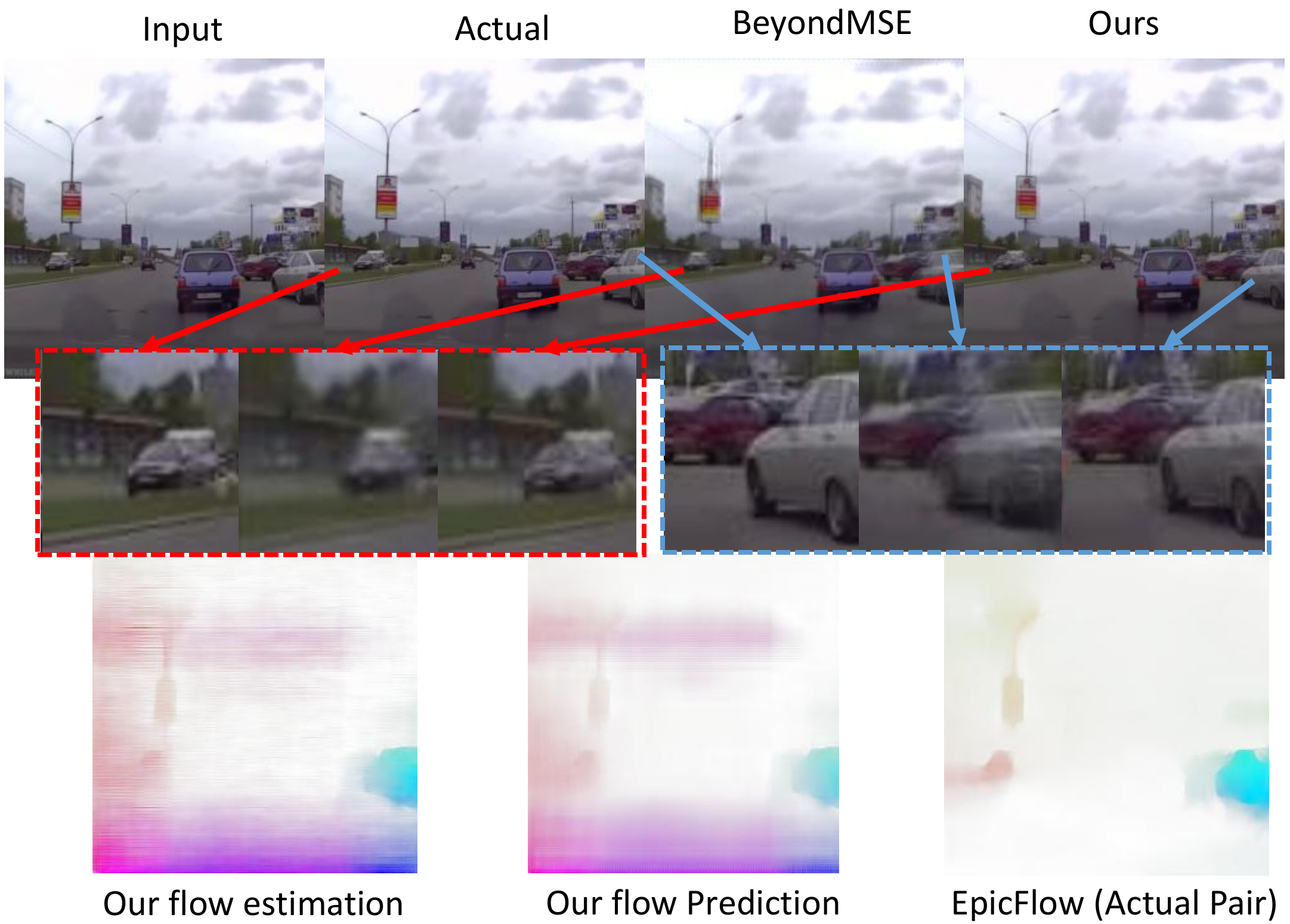}
			\caption{Qualitative results on a YouTube clip. We highlight the predicted regions of two vehicles approaching in opposite directions in red and blue boxes for better comparison.} 
			\label{fig:Youtube}
		\end{center}
		\vspace{-10mm}
	\end{figure}
	\begin{figure*}[!tp]
		\begin{center}
			\includegraphics[scale=0.58]{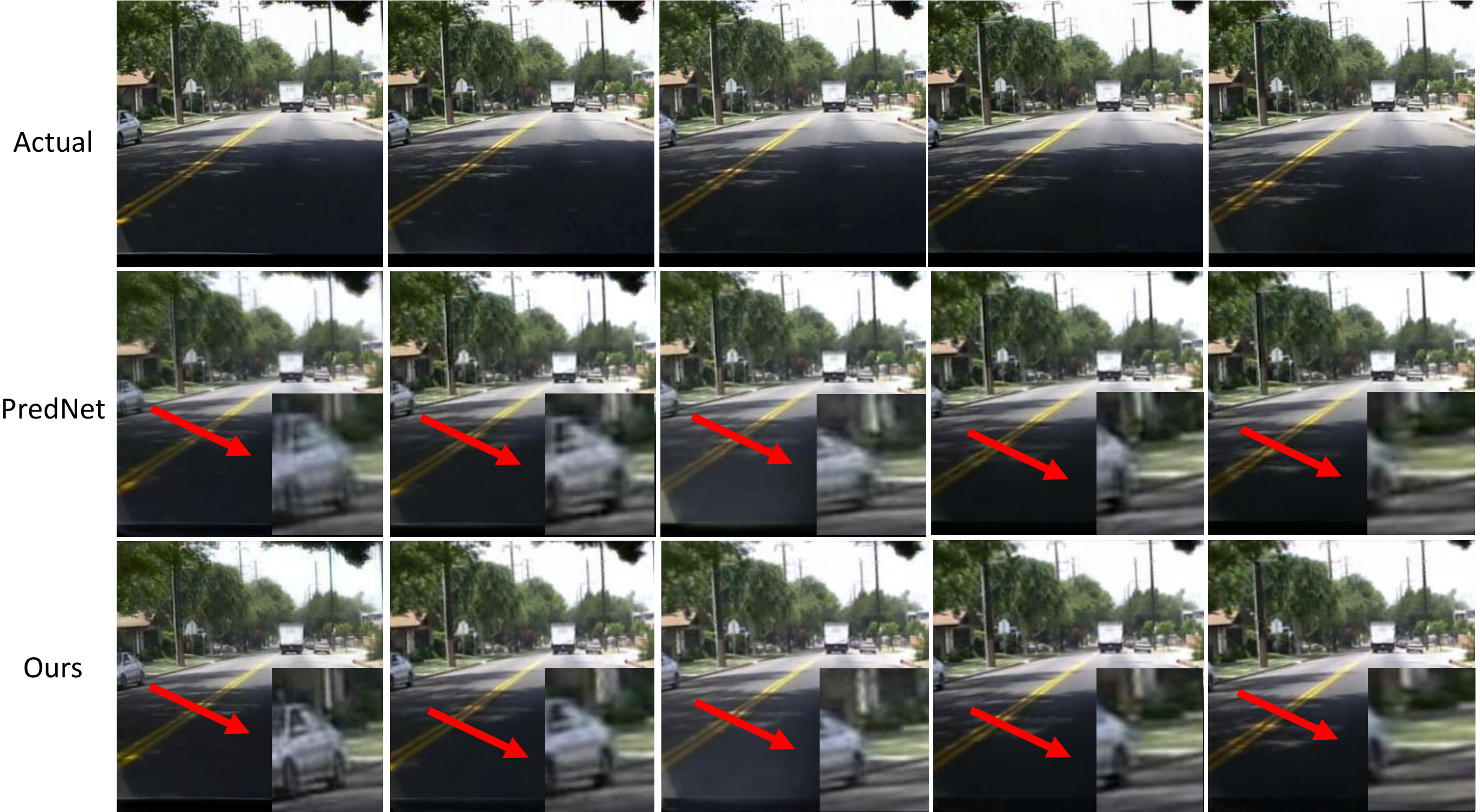}
			\caption{Qualitative comparisons with Prednet~\cite{lotter2016deep} for next-frame prediction on car-cam videos from the Caltech dataset. } 
			\label{fig:results}
		\end{center}
		\vspace{-8mm}
	\end{figure*}	
	
	\begin{figure}[!tp]
		\begin{center}
			\includegraphics[scale = 0.4]{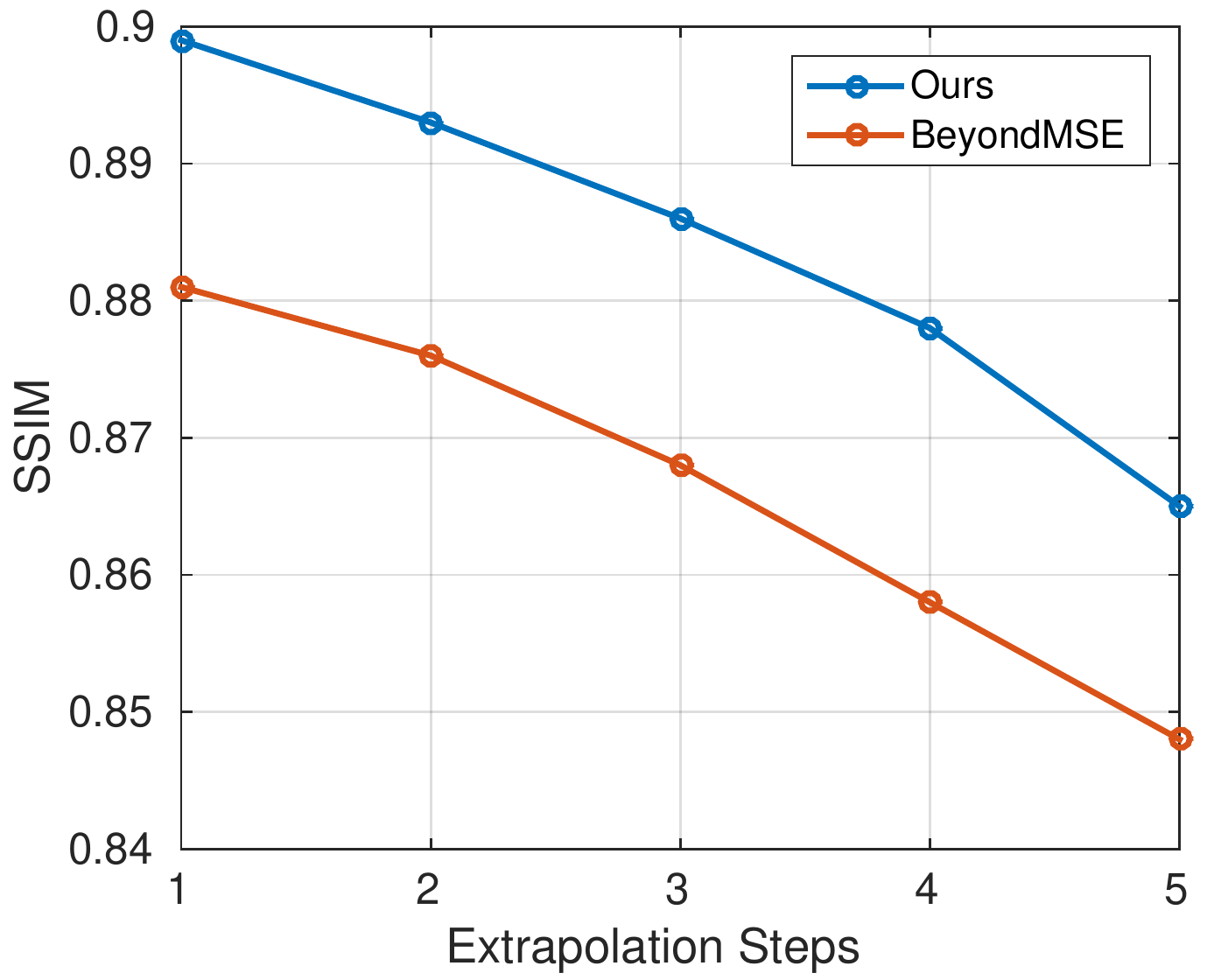}
			\caption{Performance comparison for multiple-frame prediction on the Caltech dataset.} 
			\label{fig:multistep}
		\end{center}
		\vspace{-8mm}
	\end{figure}
	
	\begin{figure*}[!tp]
		\begin{center}
			\includegraphics[scale=0.65]{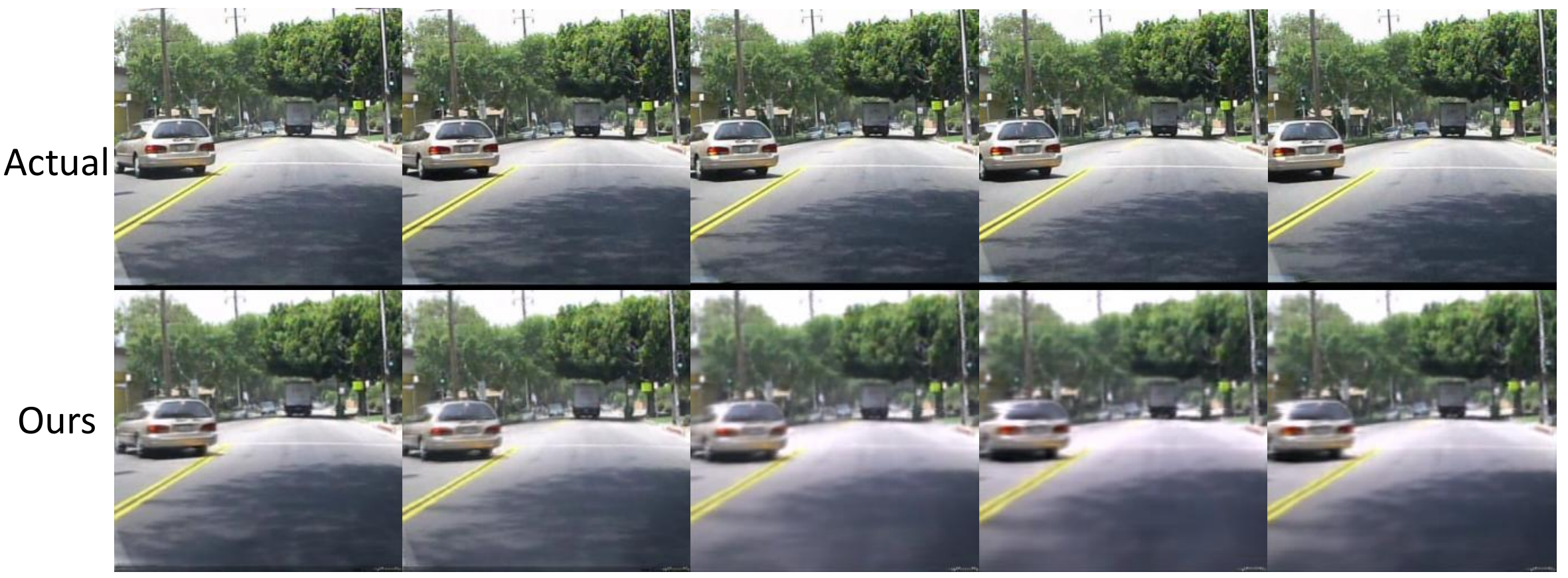}
			\caption{Multiple-frame prediction results of our model on Caltech sequences for five time steps.} 
			\label{fig:recursive}
		\end{center}
		\vspace{-7mm}
	\end{figure*}
	
	\textbf{Comparison on Caltech and YouTube Clips.} Table~\ref{tab:kitti} reports the quantitative comparison with the state-of-the-art models  BeyondMSE~\cite{mathieu2015deep} and Prednet~\cite{lotter2016deep} on the video next-frame prediction task. We obtain the results of BeyondMSE~\cite{mathieu2015deep} by training a model that minimizes the loss functions in BeyondMSE~\cite{mathieu2015deep} (ADV+GDL), and replaces the backbone network with our frame generator, except for the motion autoencoder. Our model significantly outperforms both baselines, achieving a MSE of $2.41\times10^{−3}$ and SSIM of $0.899$, compared to $3.13\times10^{−3}$ and $0.884$ of Prednet~\cite{lotter2016deep}, and $3.26\times10^{−3}$ and $0.881$ of BeyondMSE~\cite{mathieu2015deep}.
	
	We show qualitative comparisons on the Caltech Pedestrian dataset and YouTube clips in Figure~\ref{fig:results} and Figure~\ref{fig:Youtube}, respectively. In Figure~\ref{fig:Youtube}, the model is able to predict the motions of two vehicles and their shadows as they approach from different directions, as well as handle the stationary vehicle. We also show the future-flow prediction and the estimated flow between the input frame and predicted frame. Our model gives reasonable and comparable future flows with that of Epicflow~\cite{revaud2015epicflow} estimated between the actual frame pair. 
	
	\textbf{Comparison on UCF-101 and THUMOS-15.}
	Table~\ref{tab:ucf} shows the comparisons of four state-of-the-art methods on UCF-101~\cite{soomro2012ucf101} and THUMOS-15~\cite{gorban2015thumos}. We directly compare the results reported in DVF~\cite{liu2017video}. BeyondMSE~\cite{mathieu2015deep} directly hallucinates pixel values while EpicFlow~\cite{revaud2015epicflow}, DVF~\cite{liu2017video}, and Nextflow~\cite{sedaghat2016next} extrapolate future frames by predicting intermediate flows. Our dual motion GAN, which combines the merits of both frame-based and flow-based models via a dual-learning mechanism, achieves the best performance.
	
	\textbf{Multiple frame prediction.} We report the performance comparison with BeyondMSE~\cite{mathieu2015deep} in Figure~\ref{fig:multistep} and qualitative results of our model in Figure~\ref{fig:recursive}. Our model again shows better performance after five time steps, benefiting from the long-term memorization capabilities of the recurrent motion encoder.
	
	\begin{figure*}[!tp]
		\begin{center}
			\includegraphics[scale=0.6]{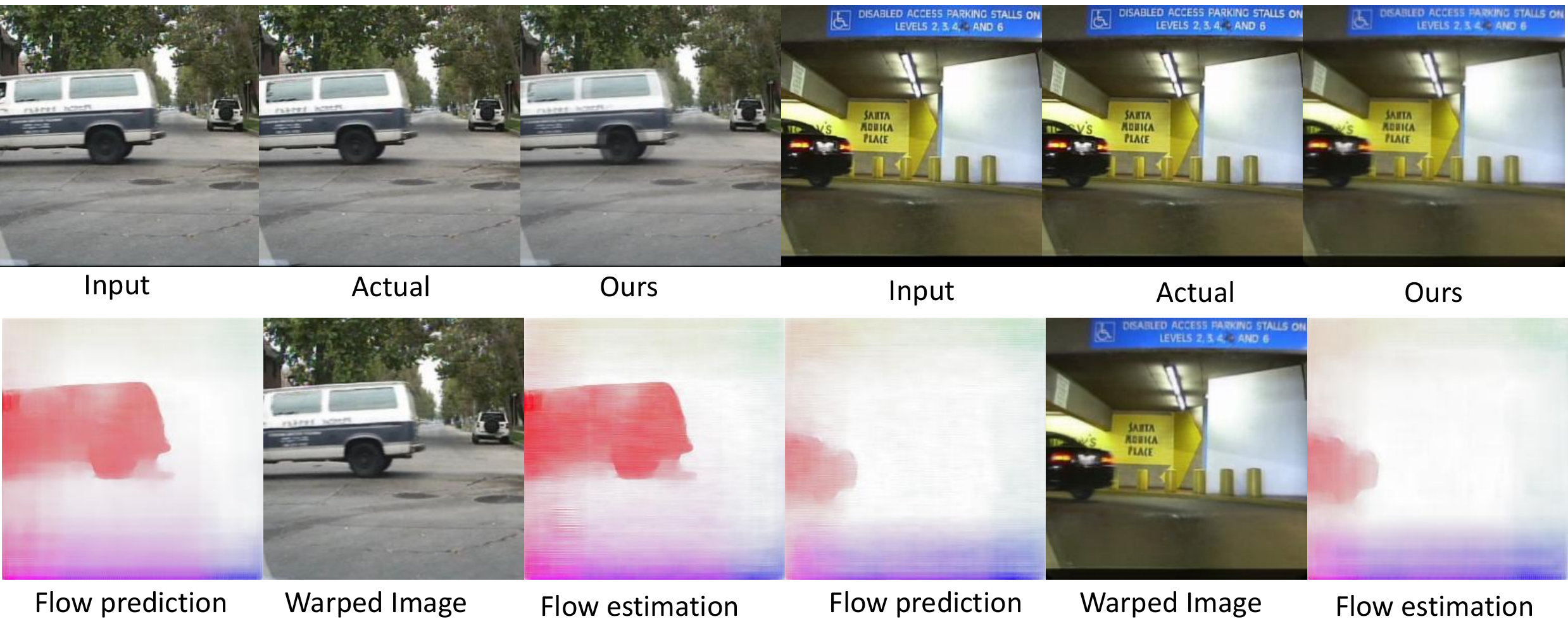}
			\caption{Some example future-frame prediction and future-flow prediction results of our model on two sequences from the KITTI dataset.} 
			\label{fig:flow}
		\end{center}
		\vspace{-7mm}
	\end{figure*}
	
	\subsection{Ablation studies}
	
	We report comparisons on our model variants in Table~\ref{tab:kitti}.
	
	\textbf{Future-frame Generator.} ``Ours frame w/o GAN'' only includes the probabilistic motion encoder and future-frame generator, and thus the final prediction is $\ddot{I}_{t+1}$. Benefiting from the motion uncertainty modeling by the probabilistic motion encoder, ``Ours frame w/o GAN'' with L1 loss achieves better results than BeyondMSE~\cite{mathieu2015deep} with L1, GDL, and ADV losses. Adding the frame discriminator and jointly optimizing the adversarial loss gives us another variant ``Ours frame GAN'', which generates sharper and more realistic predictions, and slightly outperforms PredNet~\cite{lotter2016deep}.
	
	\textbf{Future-flow Generator.} The future-flow generator learns to predict optical flows $\bar{F}_{t+1}$, which are then passed through a warping layer to get the final frame prediction $\bar{I}_{t+1}$. ``Ours flow w/o GAN'' and ``Ours flow GAN'' both obtain better prediction performance than ``Ours frame w/o GAN'' and ``Ours frame GAN'', which speaks to the superiority of learning motion fields that are more consistent with the natural characteristics of the videos. 
	
	\textbf{Adversarial Dual Objective.} We verify the advantages of combining the merits of both the future-frame generator and the future-flow generator. ``Ours frame+flow w/o GAN'' obtains remarkably lower MSE and higher SSIM over the single objective-based models ``Ours flow GAN'' and ``Our frame GAN''. Our full model has the best performance since the adversarial discriminators help judge the fidelity of both frame and flow predictions in a dual manner.
	
	\textbf{Probabilistic Motion Encoder.} We also evaluate the effect of eliminating the motion encoder (``Ours w/o motion encoder''). The significant performance drop compared to our full model can be observed, particularly on the YouTube clips. A possible reason is that the raw YouTube videos have more diverse motion patterns. The motion uncertainty of each object at different spatial locations can be effectively captured by the inferred probabilistic motion maps.
	
	\textbf{Performances of Two Generators During Testing.} The aforementioned studies use differently trained models for each setting. We also experiment on how the predictions from two generators differ from each other during testing. ``Ours future-flow (testing)'' and ``Ours future-frame (testing)'' both show significantly better results than ``Ours frame GAN'' and ``Ours flow GAN'' due to the trained dual model that can mutually improve the frame and flow predictions. We find that fusing the two predictions from the two generators obtains the best results, as shown in Figure~\ref{fig:flow}.
	
	\subsection{Flow Prediction and Estimation}
	Although we have already verified the effectiveness of flow prediction on improving future-frame prediction,
	we further quantitatively evaluate the ``by-product'' flow predictions and flow estimation performance. We compare our models with state-of-the-art models~\cite{brox2011large,revaud2015epicflow,fischer2015flownet}. Following DVF~\cite{liu2017video}, we train on the UCF-101 dataset and evaluate on the KITTI flow 2012 dataset~\cite{geiger2013vision}. The future-flow prediction module generates the predicted flows of test frames given previous frames, while the flow estimation module takes the true previous frame and test frame as inputs to estimate intermediate flows. Table~\ref{tab:flow} reports the average endpoint error (EPE) over all the labeled pixels. Our dual motion GAN only uses the flows predicted by EpicFlow~\cite{revaud2015epicflow} as the supervision for training the flow estimator and future-flow prediction module, which is thus an unsupervised method. Both the performances of our flow prediction and flow estimation are competitive with existing methods. ``Ours (flow GAN)'', in which future-frame prediction is eliminated during training, is inferior to our full model ``Ours (flow prediction)'', but is better than the prior flow-based method DVF~\cite{liu2017video}. Our model is capable of encoding essential motion information, benefiting from the joint optimization of the primal video frame prediction and dual flow prediction tasks. In addition, forecasting future flows is more challenging than flow estimation given two true frames. We provide visualization results by flow prediction and estimation in Figure~\ref{fig:flow}.
	
	\begin{table}[!tp]
		\centering\renewcommand\arraystretch{1.1}\footnotesize
		\caption{Endpoint error of flow estimation and prediction on the KITTI dataset. Here, lower values indicate better performance. }\label{tab:flow}
		\begin{tabular}{c|cccccccccccccccccccccp{6em}p{5em}}
			\toprule
			Method & EPE\\
			\hline
			Flownet~\cite{fischer2015flownet}  (supervised) & 9.1\\
			\hline
			EpicFlow~\cite{revaud2015epicflow} (unsupervised) & 3.8\\
			\hline
			DVF~\cite{liu2017video} (unsupervised) & 9.5\\
			\hline
			{Ours (flow GAN)} (unsupervised) & {9.3}\\
			\hline
			\textbf{Ours (flow prediction)} (unsupervised) & \textbf{8.9}\\
			\textbf{Ours (flow estimation)} (unsupervised) & \textbf{7.6}\\
			\bottomrule
		\end{tabular}%
		\vspace{-7mm}
	\end{table}%
	
	\subsection{Unsupervised Representation Learning}
	
	To show the effectiveness of our model on unsupervised video representation learning, we replace the future-frame and future-flow generators with one fully-connected layer and one softmax loss layer appended to the probabilistic motion encoder. Our model is then fine-tuned and tested with an action recognition loss on the UCF-101 dataset (split-1), following~\cite{mathieu2015deep,liu2017video}. This is equivalent to treating the future-frame and future-flow prediction tasks as pre-training. As demonstrated in Table~\ref{tab:action}, our model outperforms random initialization by a large margin and also shows superior performance compared to other approaches.
	
	\begin{table}[!tp]
		\centering\renewcommand\arraystretch{1.1}\footnotesize
		\caption{Classification accuracy of action recognition on UCF-101.}\label{tab:action}
		\begin{tabular}{c|cccccccccccccccccccccp{6em}p{5em}}
			\toprule
			Method & Accuracy\\
			\hline
			Unsupervised Video~\cite{wang2015unsupervised} & 43.8\\
			\hline
			Shuffle\&Learn~\cite{misra2016shuffle} & 50.2\\
			\hline
			DVF~\cite{liu2017video} & 52.4\\
			\hline
			\textbf{Ours} & \textbf{55.1}\\
			\bottomrule
		\end{tabular}%
		\vspace{-8mm}
	\end{table}%
	
	\section{Conclusion and Future Work}
	
We proposed a dual motion GAN that simultaneously solves the primal future-frame prediction and future-flow prediction tasks via a dual adversarial training mechanism. The probabilistic motion encoder learns to capture spatial motion uncertainty, while the dual adversarial discriminators and generators send feedback signals to each other to generate realistic flows and frames that are implicitly coherent with each other. Extensive experiments on video frame prediction, flow prediction, and unsupervised video representation learning demonstrate the contributions of our model to motion encoding and predictive learning. As future work, we plan to explicitly model the multi-agent dependencies so as to be able to handle real-world videos with complex motion interactions.

\section*{Acknowledgement}
This work is funded by the Department of Defense under Contract No. FA8702-15-D-0002 with Carnegie Mellon University for the operation of the Software Engineering Institute, a federally funded research and development center.

	{\small
		\bibliographystyle{ieee}
		\bibliography{egbib}
	}
	
\end{document}